\documentclass[letterpaper, 10 pt, journal, twoside]{IEEEtran}

\usepackage{cite}
\usepackage{amsmath,amssymb,amsfonts}
\usepackage{algorithmic}
\usepackage{graphicx}
\usepackage{textcomp}
\usepackage{multirow}
\usepackage{xcolor}
\usepackage{leftidx}
\usepackage{color,soul}
\usepackage{caption}
\usepackage{subcaption}
\usepackage{makecell}
\usepackage[export]{adjustbox}
\usepackage{hyperref}
\usepackage{booktabs}
\usepackage[nolist]{acronym}
\usepackage{tikz}
\usepackage{tabularx}
\usepackage{makecell}
\usepackage{pict2e}

\let\labelindent\relax
\usepackage[inline]{enumitem}

\newacro{slam}[SLAM]{Simultaneous Localization and Mapping}
\newacro{uav}[UAV]{Unmanned Aerial Vehicle}
\acrodefplural{uav}[UAVs]{unmanned aerial vehicles}
\newacro{gns}[GNS]{Global Navigation Satellite}
\newacro{gnss}[GNSS]{Global Navigation Satellite System}
\acrodefplural{gnss}[GNSS]{Global Navigation Satellite Systems}
\newacro{mcl}[MCL]{Monte-Carlo localization}
\newacro{imu}[IMU]{Inertial Measurement Unit}
\newacro{dof}[DOF]{degree-of-freedom}
\acrodefplural{dof}[DOFs]{degrees-of-freedom}
\newacro{ransac}[RANSAC]{random sample consensus}
\newacro{map}[MAP]{maximum a posteriori}
\newacro{mle}[MLE]{maximum likelihood estimation}
\newacro{rms}[RMS]{root-mean-square}
\newacro{dem}[DEM]{digital elevation model}
\newacro{vio}[VIO]{visual-inertial odometry}
\newacro{cnn}[CNN]{convolutional neural network}
\newacro{pdf}[pdf]{probability density function}
\newacro{mi}[MI]{mutual information}
\newacro{ssd}[SSD]{sum of squared differences}
\newacro{vpr}[VPR]{visual place recognition}
\newcommand{\figref}[1]{\hyperref[#1]{Fig.~\ref*{#1}}}
\newcommand{\tabref}[1]{\hyperref[#1]{Tab.~\ref*{#1}}}
\newcommand{\secref}[1]{\hyperref[#1]{Sec.~\ref*{#1}}}
\newcommand{\algoref}[1]{\hyperref[#1]{Alg.~\ref*{#1}}}

\newcommand{\cond}[2]{p(#1\vert#2)}

\def\xycoords{$(x, y)$-coordinates}

\def\ground{ground-truth}

\def\ie{\textit{i.e.},}
\def\eg{\textit{e.g.},}
\def\etal{\textit{et al.}}

\def\figvspace{\vspace{-.4em}}

\def\numrefmethods{six}

\newcommand{\visualizeuavsample}[2]{%
    \rotatebox{90}{\parbox[c]{2.5cm}{\centering \scriptsize #2}} &
    \includegraphics[width=0.135\linewidth]{images/uavvisualizations/#1_uav.pdf} &
    \includegraphics[width=0.1\linewidth]{images/uavvisualizations/#1_measurement.pdf} &
    \includegraphics[width=0.1\linewidth]{images/uavvisualizations/#1_mapcrop.pdf} &
    \includegraphics[width=0.1\linewidth]{images/uavvisualizations/#1_mapvisualization.pdf} &
    \includegraphics[width=0.135\linewidth]{images/uavvisualizations/#1_translation.pdf} &
    \includegraphics[width=0.1\linewidth]{images/uavvisualizations/#1_rotation.pdf} &
    \includegraphics[width=0.1\linewidth]{images/uavvisualizations/#1_scale.pdf} \\  
}

\newcommand{\visualizeorthomap}[1]{%
    \rotatebox[origin=t]{90}{\parbox[c]{1.2cm}{\centering \small #1}} &
    \includegraphics[width=\linewidth]{images/orthophotos/#1_map.jpg} &
    \includegraphics[width=\linewidth]{images/orthophotos/#1_summer_to_summer_measurements.jpg} &
    \includegraphics[width=\linewidth]{images/orthophotos/#1_winter_to_summer_measurements.jpg} \\
}

\title{
Season-invariant GNSS-denied visual localization for UAVs
}

\author{Jouko Kinnari$^{1}$, Francesco Verdoja$^{2}$ and Ville Kyrki$^{2}$
\thanks{Manuscript received: February 22, 2022; Revised: June 2, 2022; Accepted: July 6, 2022.}
\thanks{This paper was recommended for publication by
Editor Pauline Pounds upon evaluation of the Associate Editor and Reviewers’
comments.}
\thanks{This work was supported by Saab Finland Oy.}
\thanks{$^{1}$J. Kinnari is with Saab Finland Oy,
Salomonkatu 17B, 00100 Helsinki, Finland
{\tt\small jouko.kinnari@saabgroup.com}}%
\thanks{$^{2}$F. Verdoja and V. Kyrki are with School of Electrical Engineering, Aalto University, Finland {\tt\small \{firstname.lastname\}@aalto.fi}}%
}

\begin{document}

\maketitle

\begin{abstract}
Localization without \acp{gnss} is a critical functionality in autonomous operations of \acp{uav}. Vision-based localization on a known map can be an effective solution, but it is burdened by two main problems: places have different appearance depending on weather and season, and the perspective discrepancy between the \ac{uav} camera image and the map make matching hard. In this work, we propose a localization solution relying on matching of \ac{uav} camera images to georeferenced orthophotos with a trained \acl{cnn} model that is invariant to significant seasonal appearance difference (winter-summer) between the camera image and map. We compare the convergence speed and localization accuracy of our solution to \numrefmethods{} reference methods. The results show major improvements with respect to reference methods, especially under high seasonal variation. We finally demonstrate the ability of the method to successfully localize a real \ac{uav}, showing that the proposed method is robust to perspective changes.
\end{abstract}

\acresetall

\section{Introduction}

Knowing the Earth-fixed coordinates of an \ac{uav} is one of the basic functionalities required for long-distance autonomous \ac{uav} flight. Traditionally, \acp{gnss} have been used. However, \acp{gnss} are vulnerable to intentional jamming and spoofing attacks by an adversary, and naturally susceptible to blockages and reflections in radio signal paths.

In an ideal localization system, the \ac{uav} could infer its location using onboard sensors, without having to depend on availability of any infrastructure. One viable sensor set is a combination of \ac{imu} and camera. Inertial and \ac{vio} solutions \cite{Scaramuzza2020} provide tracking for the egomotion of the vehicle in the short term. As these solutions integrate noisy signals, a significant localization error accumulates over a longer period without further correction. \ac{slam} \cite{7747236} systems help in reducing this error in case the \ac{uav} traverses the same area a number of times during a mission. However, the correction of accumulated drift provided by \ac{slam} is only partial, and neither \ac{slam} nor \ac{vio} systems provide georeferenced coordinates without additional information.

In order to provide georeferenced coordinates, a way of matching sensor observations against a georeferenced map is needed. Providing a match between the observations of the \ac{uav} and a map is, however, not a trivial task. Not only do imaging conditions vary due to differences in the imaging hardware, illumination, and weather, but the appearance of the environment changes significantly over seasons.

\begin{figure}
    \centering
    \includegraphics[width=.9\linewidth]{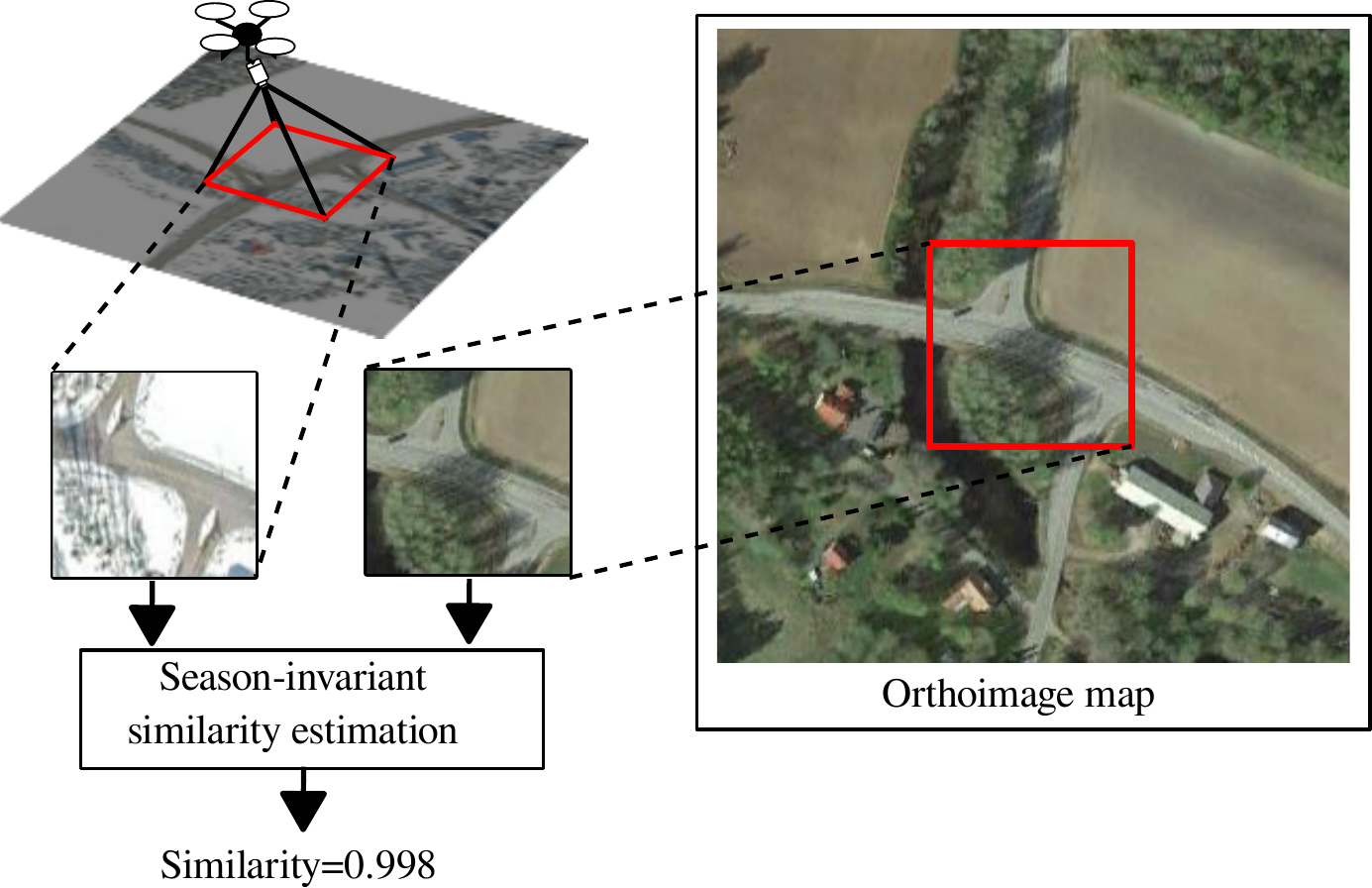}
    \caption{We develop a similarity scoring method for \ac{uav} localization that is invariant to seasonal appearance change.}
    \label{fig:first_page_figure}
    \figvspace{}
\end{figure}

We propose an image matching approach to season-invariant localization, where the information contained in an image acquired by a \ac{uav} is used for verifying or disputing correspondence to an orthoimage map. Using satellite image data, we train a model to learn a similarity measure between orthoimages in a way that is invariant to seasonal change (\figref{fig:first_page_figure}), and utilize that model for \ac{uav} localization in a \ac{mcl} \cite{ThrunProbabilisticRobotics05} framework. We demonstrate that, starting from imprecise initialization, the presented method provides significantly shorter time to convergence and smaller localization error than \numrefmethods{} baseline methods. Moreover, we illustrate the method's operation with real-world data from three \ac{uav} flights.

The main contributions of this paper are (i) a solution to visual UAV geolocalization over significant seasonal variation using a Siamese \ac{cnn}, (ii) a method using Gaussian kernel density estimation to evaluate the confidence of the \ac{cnn} output to be used with \ac{mcl}, and (iii) a demonstration of the robustness of the solution using real-world and simulation data of flights under significant seasonal change.

\section{Related work}
\label{sec:related_work}

A key functionality in image-based \ac{uav} localization is providing a way to find the correspondence between the \ac{uav} image and a map. Classical manually engineered feature detectors and descriptors such as SIFT \cite{lowe2004distinctive} have been proposed by \eg{} Cesetti \etal{} \cite{Cesetti2011AVG}, but large changes in perspective and seasonal appearance pose challenges in finding correspondences. Feature descriptors specifically hand-crafted for \ac{uav} localization have also been proposed by Mantelli \etal \cite{MANTELLI2019304}, who modified the BRIEF descriptor \cite{brief} to utilize color information. However, we expect color information not to be reliable over significant seasonal appearance change.

Learned features may provide more robust observations for localization. An example of a recent deep learning-based feature detector and descriptor applied to \ac{uav} localization is by Hou \etal{} \cite{9311612}, who demonstrate reduction of odometry error of a \ac{uav} in short trajectories (750 m) in presence of seasonal appearance change. The approach is based on minimizing reprojection errors of a combination of D2-Net \cite{Dusmanu_2019_CVPR} features for map matching and ORB \cite{6126544} features for visual odometry. The proposed bundle adjustment approach requires accurate knowledge of initial pose and authors show that the solution is not robust to long intervals between keyframes containing map matching features. The need for accurate initialization and good image-to-map feature point matches at short intervals are significant downsides in \ac{uav} localization when operating over terrains with long periods of natural ambiguity (\eg{} lakes, fields).

Semantic features have been used for finding correspondences between \ac{uav} images and a map. In several works, the observed \ac{uav} image is first translated into an intermediate terrain class classification (using a single class such as buildings \cite{Choi2020BRMLU} or roads \cite{Dumble2015AirborneVN, volkova2018more} or multiple classes \cite{Schleiss2019TRANSLATINGAI, masselli2016, 8575361}). Next, features in the semantic representation (such as road \cite{volkova2018more} and intersection geometry \cite{Dumble2015AirborneVN,  volkova2018more}, generic shape descriptors \cite{8575361} or ORB \cite{6126544} features detected on segmented images \cite{masselli2016}) are used as landmarks for localization. Template matching-based methods include computing \ac{ssd} of semantic classes between map and \ac{uav} image \cite{Schleiss2019TRANSLATINGAI} and computing the ratio of building to non-building pixels as a matchable descriptor \cite{Choi2020BRMLU}.

Instead of matching engineered features, image-to-map matching can be performed in a latent space. Samano \etal{} \cite{9562005} recently proposed to match \ac{uav} images to map by using low-dimensional (16D) embeddings. The projection is learned by finding compatible embeddings for a \ac{uav} image and a corresponding semantic map. The embeddings are matched using Euclidean distance. Successful localization is demonstrated for simulated \ac{uav} images using \ac{mcl}.
The high reported matching performance and the availability of source code make \cite{9562005} a good comparison approach for our method. Couturier \etal{} \cite{couturier2021} proposed a similar architecture for global descriptor vector extraction.

Instead of detecting visual feature points or relying in any way on a semantic representation, it is possible to use the full image area for finding the correspondence of the \ac{uav} image and an orthophoto map. Recent template matching approaches include the use of Pearson correlation \cite{pearson1896vii} in assessing top-down \ac{uav} image similarity to an orthoimage map \cite{Jureviius2019RobustGL}. In \cite{Patel2020VisualLW, 9357892}, authors match \ac{uav} images to precomputed images rendered from preplanned flight paths. To target generality, our focus is on deriving a solution in which the planned path of the autonomous agent is allowed to change during a mission, without requiring significant computation before starting to follow the plan.

Another way to approach the \ac{uav} localization problem is to consider it as a sequence of homography transformations between the \ac{uav} image and a base map. Yol \etal{} \cite{6943040} vary homography parameters and maximize \ac{mi}. Goforth \etal{} \cite{8793558} take a similar approach but add a learning model that transforms the original camera image to a learned feature space to gain a level of seasonal invariance. In both works, starting pose is assumed known. Both solutions track a single hypothesis, which is likely to lead to loss of tracking capability in case of a long period of ambiguity in terrain or due to intermittent matching errors.

To enable tracking of multiple pose hypotheses across ambiguous regions, our work and multiple others (\eg{} \cite{Jureviius2019RobustGL, masselli2016, 9562005, MANTELLI2019304}) use \ac{mcl} \cite{ThrunProbabilisticRobotics05} to fuse odometry and map observations.

Outside \ac{uav} localization literature, seasonal variations have also been addressed in  the context of \ac{vpr} \cite{9336674}, serving as inspiration for our work.

In contrast to the works mentioned above, we present a way to measure the correctness of a pose hypothesis without relying on human-chosen features or semantics. 
We expect this design choice to allow the matching method to learn a meaningful representation without being constrained to an explicit definition of semantic terrain classes, and without being dependent on existence of those classes in the terrain over which the flight occurs.

\section{Methodology}
\label{sec:methodology}

\subsection{\ac{uav} Localization}
\label{ssec:proposed_localization_solution}

In the full localization problem, there are 6 degrees of freedom to estimate. To reduce the dimensionality of the problem, we assume that the roll and pitch angles of the \ac{uav} can be inferred from the direction of gravity, measurable with an \ac{imu}. Altitude is inferred as part of orthoprojection method as presented in earlier work \cite{kinnari2021gnssdenied}. The state is then defined as

\begin{equation}
X = (x,y,\phi,s)\enspace,
\end{equation}

where $x$, $y$ are longitude and latitude of the \ac{uav} position in the map coordinate system, $\phi$ is the yaw of the \ac{uav} and $s$ is a scale parameter, which allows the solution to work in case of scale drift.

To estimate $X$, we choose \ac{mcl}, a particle filter tailored for localization over a map $\mathcal{M}$ known in advance. \Ac{mcl} represents the belief on the estimated pose at time $t$ as a set of particles $X_t^r, r=1\ldots N$. For initialization, we assume a uniform distribution on an interval for values of $x$, $y$ and $s$ and uniform distribution over all values of heading for $\phi$. When an odometry measurement is available, a new pose for each particle is sampled in accordance with the distribution of the odometry measurement. A separate observation $I_t$ of the environment acquired in the new pose is then used to update the weight of each particle. The weight $w_r$ of particle $r$ is $w_r = P(X_t^r|\mathcal{I}_t,\mathcal{M})$, \ie{} the likelihood that the pose $X_t^r$ represents the true pose, given observation $\mathcal{I}_t$ and map $\mathcal{M}$. The likelihood is estimated based on a similarity measure as described in the following section. This weight is used when resampling a new representative particle set. We follow the particle filter algorithm and low variance sampler described in~\cite{ThrunProbabilisticRobotics05} and resample after each update.

\subsection{Similarity scoring function structure}
\label{ssec:similarity_scoring_function_structure}

Given a map $\mathcal{M}$, an observation by the \ac{uav}, $\mathcal{I}_t$, and a pose hypothesis $X_t^r$, we first perform an orthoprojection of the \ac{uav} image with the method presented in earlier work \cite{kinnari2021gnssdenied}.  
The orthoprojection method in \cite{kinnari2021gnssdenied} performs \ac{vio} and estimates the position of tracked \ac{vio} features (landmarks) with respect to drone coordinate frame in meters, assuming sufficient excitation on inertial measurements to resolve scale \cite{5959226}. By assuming the ground beneath the \ac{uav} is planar, the parameters of a plane that best fits the landmark coordinates are resolved and the \ac{uav} image is projected to a top-down view by planar homography. This allows creating a projection of the \ac{uav} image at a desired resolution (1 m/px in our case) independent of \ac{uav} altitude. To tolerate slight drift in estimated scale, our state estimator includes the scale parameter.

From the orthoprojected \ac{uav} image, we find the corner points of a square 96$\times$96 m area (1 m/pixel) close to nadir view that is fully visible in the \ac{uav} image. Given these corner point locations, we compute what are the corresponding points on the map, if pose hypothesis $X_t^r$ was correct. We then crop a square image patch from both the map $\mathcal{M}$ and the \ac{uav} image $\mathcal{I}_t$. We denote these image patches $I_{\mathcal{M},X_t^r}$ and $I_{t}$, respectively. Examples are visualized in \figref{fig:uav_sample_pictures}.

These image patches are used to compute a similarity score $c_t^r$, for each pose hypothesis indexed by $r$ at time $t$, using a similarity function $f$:

\begin{equation}
    c_t^r = f(I_{\mathcal{M},X_t^r},I_{t})
    \label{eq:similarity_function}
\end{equation}

\begin{figure}
    \centering
    \includegraphics[width=\linewidth]{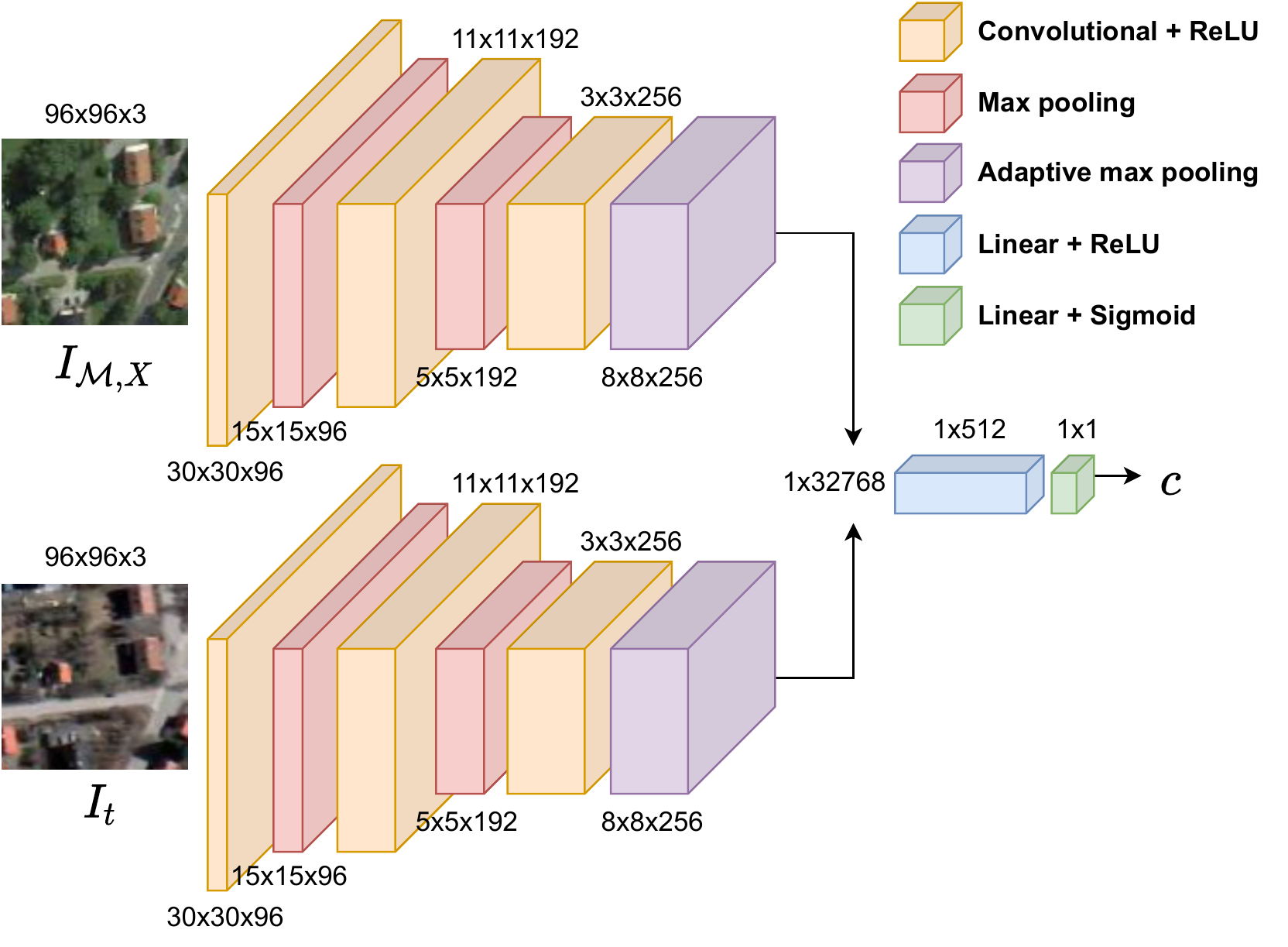}
    \caption{Model structure for the image similarity network $c = f(I_{\mathcal{M},X},I_{t})$. Branches share parameters.}
    \label{fig:modelstructure}
    \figvspace{}
\end{figure}

To learn $f$, we propose an image comparison \ac{cnn} architecture inspired by \cite{Zagoruyko_2015_CVPR}. The model structure is shown in \figref{fig:modelstructure}. The model contains two Siamese network branches and a separate decision network\footnote{Trained model and instructions for downloading training data available at \url{https://irobotics.aalto.fi/sivl}}. The model takes as input the pair of images ($I_t$, $I_{\mathcal{M},X}$) and it produces a similarity measure $c \in [0,1]$.

\subsection{Data}
\label{ssec:data}

To train the model, we use satellite images from datasets released in \cite{8793558} and collect additional data from Google Earth historical images to include seasonal variation.

There are a total of 18 different areas with 3 to 15 satellite images acquired per area. For each area, we define a grid of possible sampling locations. For each location, per each epoch, we select one random satellite image from that area, crop a 96$\times$96 m area with random yaw and small random translation around grid point, and label that sample \emph{anchor}. We also generate another sample using the same crop parameters but from another image of the same area, which we label the \emph{positive} sample. A third, \emph{negative} sample is generated by selecting a random location and orientation within the same area. The training dataset consists of 21870 unique locations for sampling, the testing dataset contains 1392 locations, and the weighing function estimation dataset has 5568 locations.

We perform various augmentations during training using \cite{2018arXiv180906839B}. We apply random flips and transposes on the data, to generate more data. We also apply Gaussian noise, various means of blur, sharpening, emboss, brightness, contrast and color changes to gain additional robustness to illumination changes and imaging noise. Additionally, to provide robustness against orthoprojection errors, we apply small geometric transformations.

In training, we use binary cross-entropy loss, setting target to $1$ for the pair of anchor and positive samples, and to $0$ for the pair of anchor and negative samples. The model is trained using Adam optimizer with learning rate $10^{-5}$ and a weight decay of $10^{-8}$ for 1000 epochs.

\subsection{Computing importance factor from similarity measure} \label{ssec:computing_likelihood}

We want to calculate the importance factor $w_t^r$ for each particle $r$ at time $t$ to incorporate the camera observation in the particle set. We compute the importance factor as
$w^t_r = P(X_t^r|\mathcal{I}_t,\mathcal{M}) = \cond{S=s}{c_t^r}$, where $S = \{s, u, o\}$ is a variable that determines if the measurement was a match ($S=s$), not a match ($S=u$), or  an outlier ($S=o$), and $c_t^r$ is determined by \eqref{eq:similarity_function}. We assume $p(o) = \beta$ where $\beta = 0.05$. As we resample after each update, we assume uninformed priors $p(s) = p(u) = (1-\beta)/2$. We calculate the importance factor as

\begin{equation} \label{eq:post}
    \cond{s}{c_t^r} = \frac{\cond{c}{s}p(s)}{\sum_{i\in \{s,u,o\}}\cond{c}{i}p(i)}
\end{equation}

\begin{figure}[t]
    \centering
    \begin{subfigure}[t]{0.45\linewidth}
        \centering
        \includegraphics[width=\linewidth]{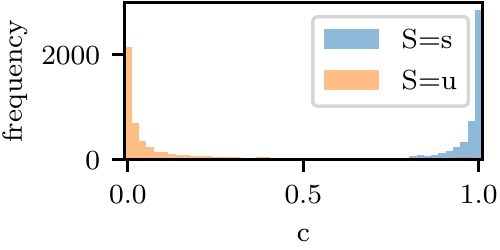}
        \caption{Histograms}
    \end{subfigure}%
    ~ 
    \begin{subfigure}[t]{0.45\linewidth}
        \centering
        \includegraphics[width=\linewidth]{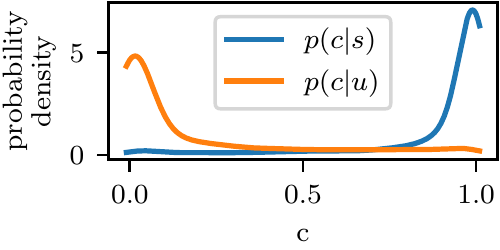}
        \caption{Probability densities}
    \end{subfigure}
    \caption{Histograms of similarity measures and estimated probability densities.}
    \label{fig:score_histogram_and_pdf}%
\end{figure}

To compute \eqref{eq:post}, we estimate the \ac{pdf} $\cond{c}{s}$ from samples corresponding with true pose, using Gaussian kernel density estimation with Scott's bandwidth rule \cite{scott1992}.
Similarly, we estimate a \ac{pdf} corresponding with incorrect pose $\cond{c}{u}$ for a number of randomly drawn poses. To collect samples corresponding with true pose ($S=s$), we extract pairs of subimages from satellite images corresponding with same area in a similar way as in \secref{ssec:data} and compute values of $c$ for each pair from \eqref{eq:similarity_function}. To collect samples corresponding with false pose ($S=u$), we extract pairs of subimages from non-corresponding locations. In estimating $\cond{c}{s}$ and $\cond{c}{u}$, we use satellite images from areas that were not used in training $f$. A histogram of similarity scores for the two classes and corresponding probability density functions are visualized in \figref{fig:score_histogram_and_pdf}. The outlier class \ac{pdf} $\cond{c}{o}$ is assumed to be uniform over the value range of $c$. The outlier class is included in order to avoid overly confident classifications in regions of $c$ where very small amount of data is available. 

\section{Experiments}
\label{sec:experiments}

We test the accuracy of the proposed localization system, using a learned similarity score, in two experiments. 

In the first experiment (\secref{ssec:experiment_with_orthoimages}) we want to  evaluate the seasonal invariance of the proposed solution. To this end, we test the ability of the model to localize through simulated flights over urban and non-urban locations in our dataset under both \emph{significant} appearance changes (\ie{} winter imagery against a map acquired in summer) as well as \emph{minor} appearance changes (\ie{} summer imagery against a map acquired in the summer but taken at a different time). We compare the localization performance using our similarity measure method to \numrefmethods{} other similarity measures.

The second experiment (\secref{ssec:experiment_with_real_uav_data}) attempts to identify how the proposed localization solution works on real \ac{uav} data, which include a perspective change as well.

\subsection{Experimental setting}
\label{ssec:common_settings}

\subsubsection{Prior on initial pose} In each experiment, the \ac{mcl} filter is initialized with 1000 particles in a 100$\times$100 m area around the true starting position, with scale $0.95 \ldots 1.05$ and with no a priori information on yaw. This represents a situation where an end user is able to state an inaccurate starting location for the flight of a \ac{uav}, without having to input information on initial orientation.

\subsubsection{Odometry noise} In all experiments, the \ac{uav} takes a sample image approximately every 100 meters. To simulate the impact of odometry noise, we add normally distributed noise with 2 m standard deviation in $x$ and $y$ translation and 1\textdegree~standard deviation in orientation with respect to the pose increment computed from \ground{}. These parameters are in line with typical performance reported for monocular visual-inertial odometry solutions in \acp{uav} \cite{8460664}. Scale noise is assumed to be zero-mean Gaussian with a standard deviation of 0.001.

\subsection{Localization under seasonal appearance variation}
\label{ssec:experiment_with_orthoimages}

\begin{figure}
\centering
\begin{tabularx}{\linewidth}{ cXXX }%
\visualizeorthomap{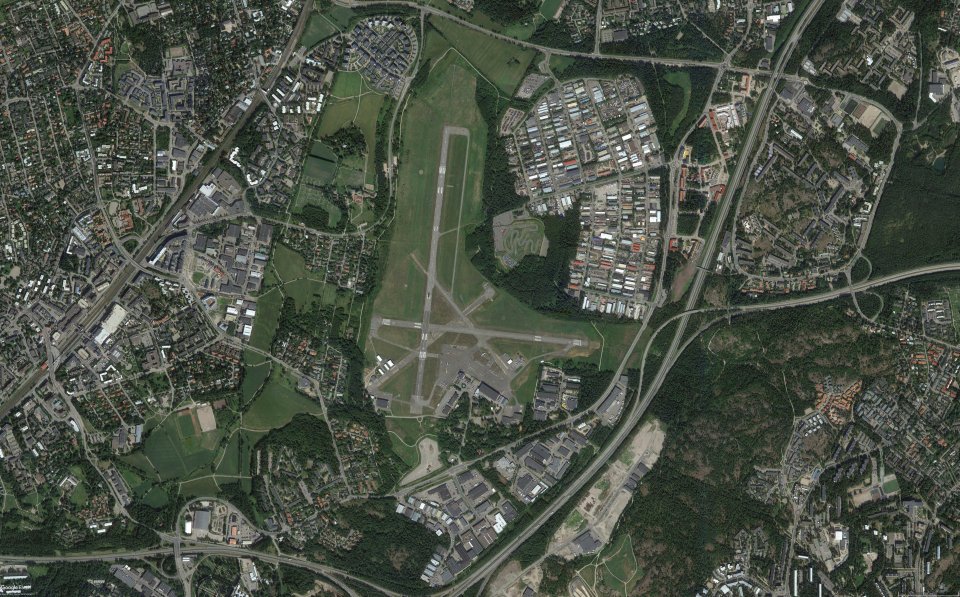}
\visualizeorthomap{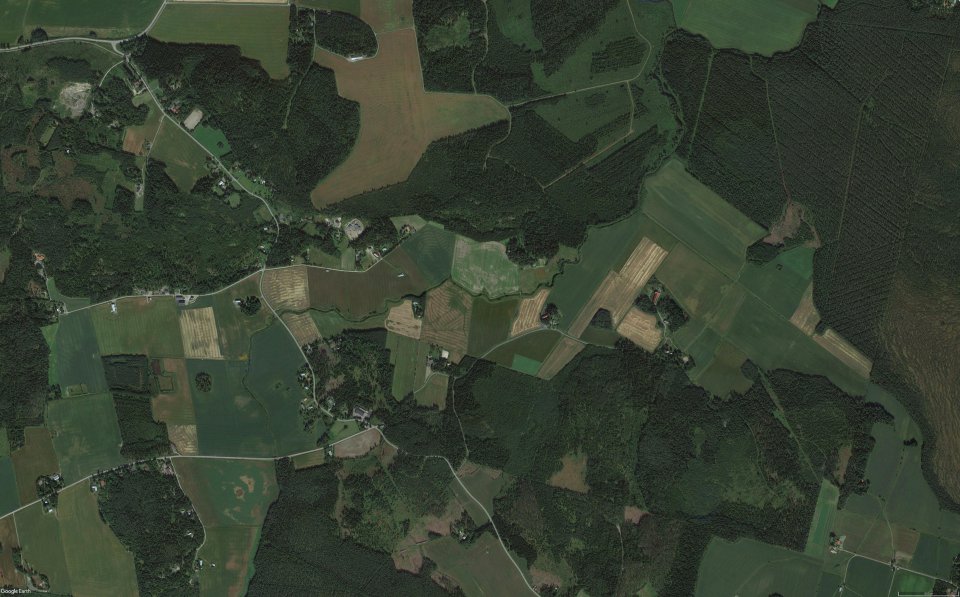}
& \centering \small (a) & \centering \small (b) & \centering \small (c) \\
\end{tabularx}
\caption{Orthoimages used in simulated experiments as map (a), ``summer'' (b), and ``winter'' (c) measurements. Each orthoimage is 4800$\times$2987 m at 1 m/pixel resolution.}
\label{fig:orthoimages}
\figvspace{}
\end{figure}

\begin{figure*}
\setlength{\unitlength}{0.1\textwidth}
\begin{picture}(10,4)
\put(0,0){\includegraphics[width=\linewidth]{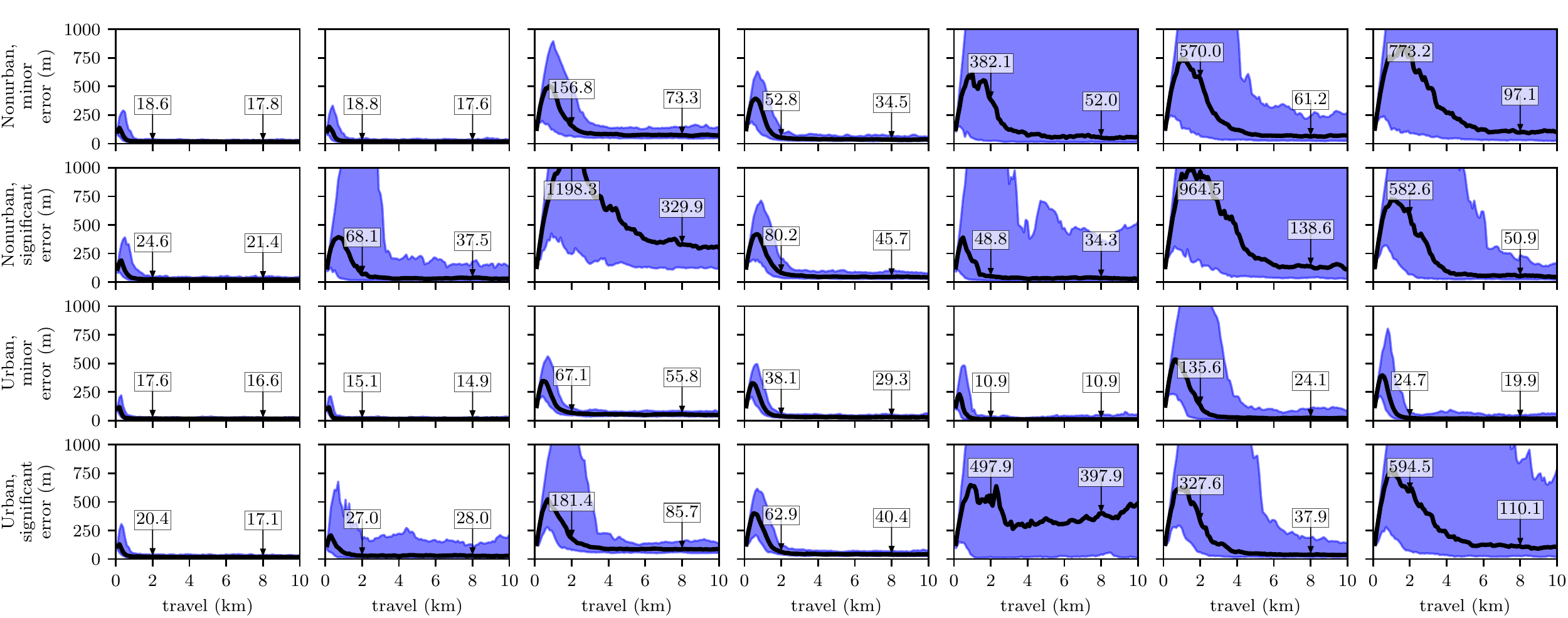}}

\put(1.200,3.84){\centering \scriptsize Ours}
\put(2.15,3.84){\centering \scriptsize Ours, w/o winter}
\put(3.45,3.84){\centering \scriptsize Samano \etal{} \cite{9562005}}
\put(4.82,3.84){\centering \scriptsize Samano retrained}
\put(6.1,3.84){\centering \scriptsize Jurevičius \etal{} \cite{Jureviius2019RobustGL}}
\put(7.95,3.84){\centering \scriptsize MI}
\put(8.800,3.84){\centering \scriptsize Kinnari \etal{} \cite{kinnari2021gnssdenied}}

\end{picture}
\caption{Errors in simulated localization experiments when using different similarity measures, over different types of terrain. \emph{Minor} appearance change refers to summer-to-summer matching, while \emph{significant} refers to winter-to-summer matching. Medians of mean errors after 20 and 80 updates annotated.}
\label{fig:ortho_errors_all}
\figvspace{}
\end{figure*}
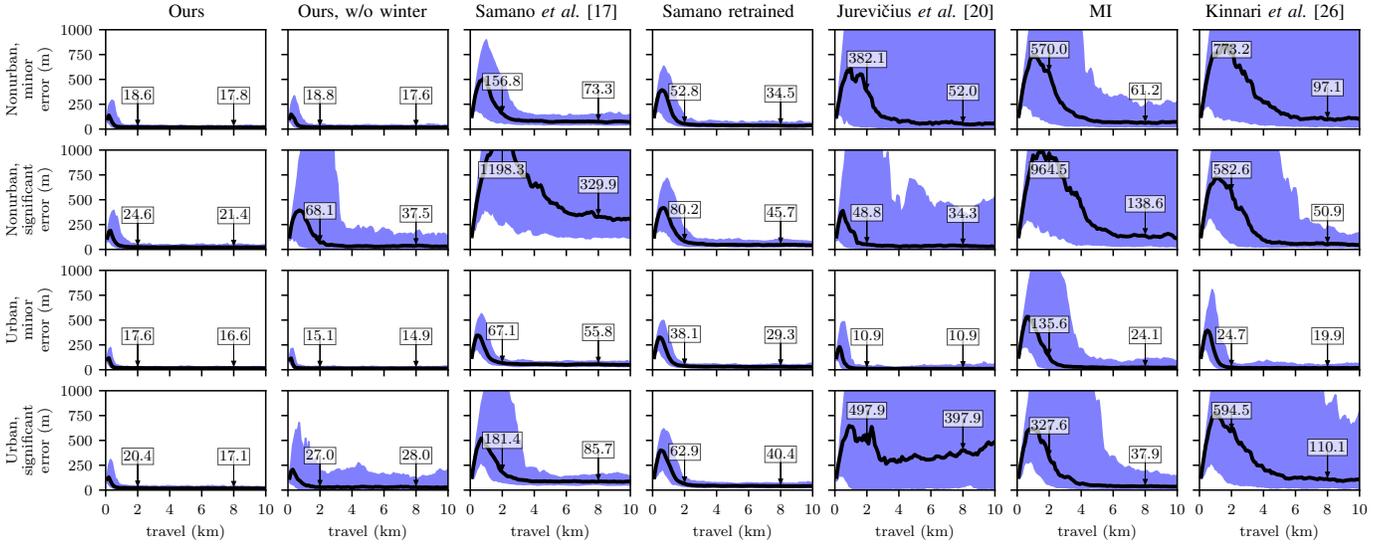

We tested the performance of the proposed localization method under minor and significant seasonal appearance change in both urban and non-urban areas. We selected two testing datasets, representing a non-urban and an urban area, respectively, selected three orthoimages from each dataset  (shown in \figref{fig:orthoimages}), and formulated an experiment where a simulated \ac{uav} flies above each area. One orthoimage acquired during summer was used as map. Another orthoimage acquired during summer was used for generating measurements for the \emph{minor} seasonal appearance case and one orthoimage acquired during winter was used for generating measurements for the \emph{significant} seasonal appearance case.

100 simulated flights were executed for each combination of minor/significant seasonal appearance and urban/non-urban area. In each run, the starting position and yaw of the simulated \ac{uav} was randomly selected within the map. Motion of the \ac{uav} was simulated with random changes in heading for a duration of 100 updates, making sure the trajectory stays within the map, and localization performance of the \ac{mcl} algorithm was quantified by computing the weighted mean Euclidean distance to \ground{} position in \xycoords{}.

We compared the localization performance achieved using the proposed method against \numrefmethods{} other map matching methods. To compare with another learning-based method, we chose Samano \etal{} \cite{9562005} that has source code and learned weights made available by the authors. To apply that approach to our problem, instead of using a semantic map, unavailable in our scenario, we leveraged their use of an intra-domain loss on \ac{uav} images in training their embedding generator network and we generated 16D embedding vectors from both $I_{\mathcal{M},X_t^r}$ and $I_t$ using their trained image feature extractor and projection modules. We computed particle weights based on distance in embedding space using the linear scaling method proposed by the authors. Performance of deep learning matching methods has rarely been evaluated in relation to more traditional metrics, which are still finding practical use in real localization applications. For this reason, we also compared against the matching method recently proposed by Jurevičius \etal{} \cite{Jureviius2019RobustGL}, using logistic conversion with $v=0.2$. As other classical image similarity measures, we used \ac{mi}, which has gained attention in other \ac{uav} localization approaches \cite{6943040, Patel2020VisualLW} and is shown by Mantelli \etal{} \cite{MANTELLI2019304} to provide marginally superior success rate to abBRIEF. Another classical similarity measure that we used as comparison is Moravec, which we evaluated in a previous work \cite{kinnari2021gnssdenied}.

In addition to Samano \etal{}, Jurevičius \etal{}, \ac{mi} and our previous approach, we also trained our model without any winter imagery (called "Ours, w/o winter" in \figref{fig:ortho_errors_all}) to ablate what we regard as the most important data augmentation source for winter-summer localization. Finally, we trained an embedding vector generator (called "Samano retrained" in \figref{fig:ortho_errors_all}) with the model structure of \cite{9562005} with the same data that we use for our proposed method. The "Samano retrained" model is trained using triplet loss, Adam optimizer, and learning rate $10^{-5}$ until the testing loss stopped improving (at approximately 200 epochs). For the "Samano retrained" model, we weighed the particles using the linear scaling method in \cite{9562005}.

In our \ac{mi} and Moravec implementations, we weighed the particles by the Gaussian kernel density estimation method described in \secref{ssec:computing_likelihood}. Each algorithm was fed the same odometry measurements and camera images, with the exception that the camera image fed to Samano \etal{}'s network was scaled to 128$\times$128 pixels and it was taken from a 95$\times$95 m area to correspond with the design choices in \cite{9562005}. Each algorithm also used the same map.

\figref{fig:ortho_errors_all} shows the median and 5\ldots95\% interval for weighted mean errors computed over all 100 runs in all permutations of urban/non-urban area type and minor/significant appearance change, using our method and the three comparison methods for similarity score computation. The initial error increase in all methods is due to no information on initial yaw. Once  particles representing false hypotheses for yaw die out, the error decreases if the correct mode in the search space is found, \ie{} if the filter converges to the correct state.

Compared to the reference methods, our method provides both faster convergence time and smaller error bounds after convergence than all the comparison algorithms. In terms of median of mean errors after convergence, Jurevičius appears to provide lower median for localization error in the case of very high texture (urban environment) and minor appearance change. We suspect this difference may be due to data augmentation by small geometric transformations that we used in training; \ie{} the network is trained to give high scores for slightly offset pairs of images. Degradation of convergence and mean error performance on our model trained without winter data show the value of training on data that contains the expected variability. The same can be seen in the comparison between Samano \etal{} network and the retrained Samano model.

\begin{figure}
    \centering
     \begin{subfigure}[t]{0.45\linewidth}
        \centering
         \includegraphics[width=\linewidth]{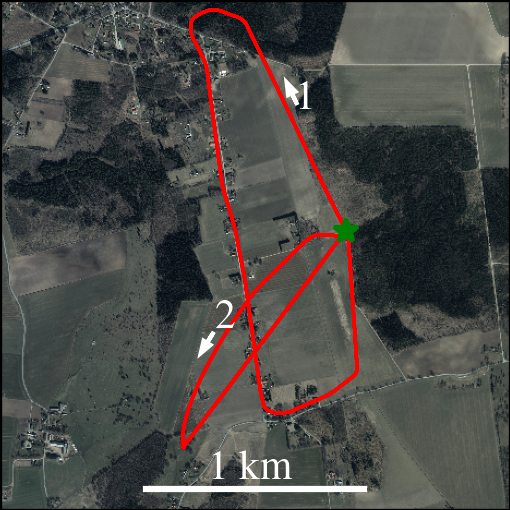}
         \caption{Map of area A (Klockrike, Sweden) and trajectory of \ac{uav} dataset 1.}
         \label{fig:area_a_map}
     \end{subfigure}~
     \begin{subfigure}[t]{0.45\linewidth}
        \centering
         \includegraphics[width=\linewidth]{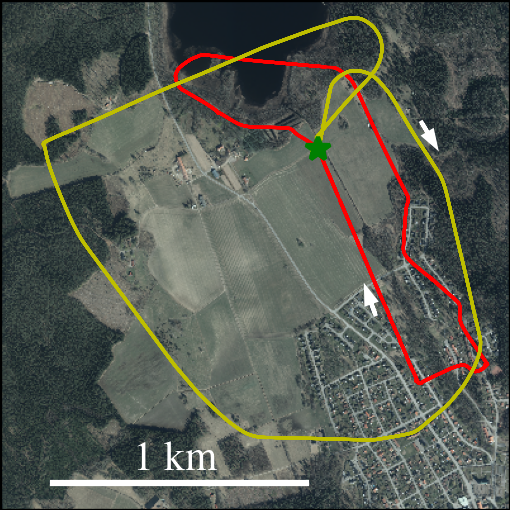}
         \caption{Map of area B (Kisa, Sweden) and trajectory of \ac{uav} datasets 2 (red), 3 (yellow)}
         \label{fig:area_b_map}
     \end{subfigure}
     \caption{\ground{} real \ac{uav} trajectories. Starting location marked with green star.}
     \label{fig:maps_and_trajectories}
     \figvspace{}
\end{figure}

\begin{table}
\caption{\label{tab:uav_dataset_characteristics}Characteristics of flights in experiments with real \ac{uav} data. Trajectory lengths computed along $(x, y)$ plane, and camera angles between nadir and camera principal axis.}
\begin{tabular}{p{3mm} p{5mm} p{12mm} p{3mm} p{24mm} p{13mm}} 
\toprule
Set & Area & Traj. length (m) & Alt. (m) & Mean camera angle [range] ($^{\circ}$) & Acquisition time\\
\midrule
1 & A & 6888 & 92 & 50.9 [48.5, 61.6] & Oct 2019 \\ 
2 & B & 4080 & 91 & 60.7 [55.2, 70.0] & Nov 2019 \\ 
3 & B & 6361 & 92 & 52.6 [48.9, 118.9] & Nov 2019 \\
\bottomrule
\end{tabular}
\figvspace{}
\end{table}

\subsection{Localization on real \ac{uav} data}
\label{ssec:experiment_with_real_uav_data}

Besides the simulated experiments using orthoimages, we ran an experiment with three datasets collected with a \ac{uav}\footnote{Data provided by Saab Dynamics Ab.} to identify performance gaps with our model trained on orthoimage data only. The trajectories are shown on a map in \figref{fig:maps_and_trajectories} and they cover forest areas, fields, residential areas, and a lake. Ground-truth pose of the \ac{uav} in these experiments has been obtained through RTK-GNSS which ensures precision in the centimeter range. Additional characteristics of these flights are listed in \tabref{tab:uav_dataset_characteristics}. The map used in these experiments was acquired during summer, and the \ac{uav} flights took place during autumn months. In \ac{uav} dataset 1, deciduous trees are showing autumn colors and in \ac{uav} datasets 2 and 3, deciduous trees have dropped leaves. 

In this work, the images acquired by the \ac{uav} were orthoprojected using \ground{} position information and a \ac{dem} of the environment where the flight takes place, but in a final use case, orthoprojection can be done \eg{} using the method presented in earlier work \cite{kinnari2021gnssdenied} or by the use of calibrated downward-facing camera and an altimeter. The choice to use \ac{dem} was made to exclude the impact of possible errors in elevation estimation. The \ac{dem} and the orthoimage map of the areas for \ac{uav} experiments were purchased from a local map information supplier\footnote{Lantmäteriet, \url{https://www.lantmateriet.se/}.}.

\begin{figure}
    \centering
    \includegraphics[width=0.95\linewidth]{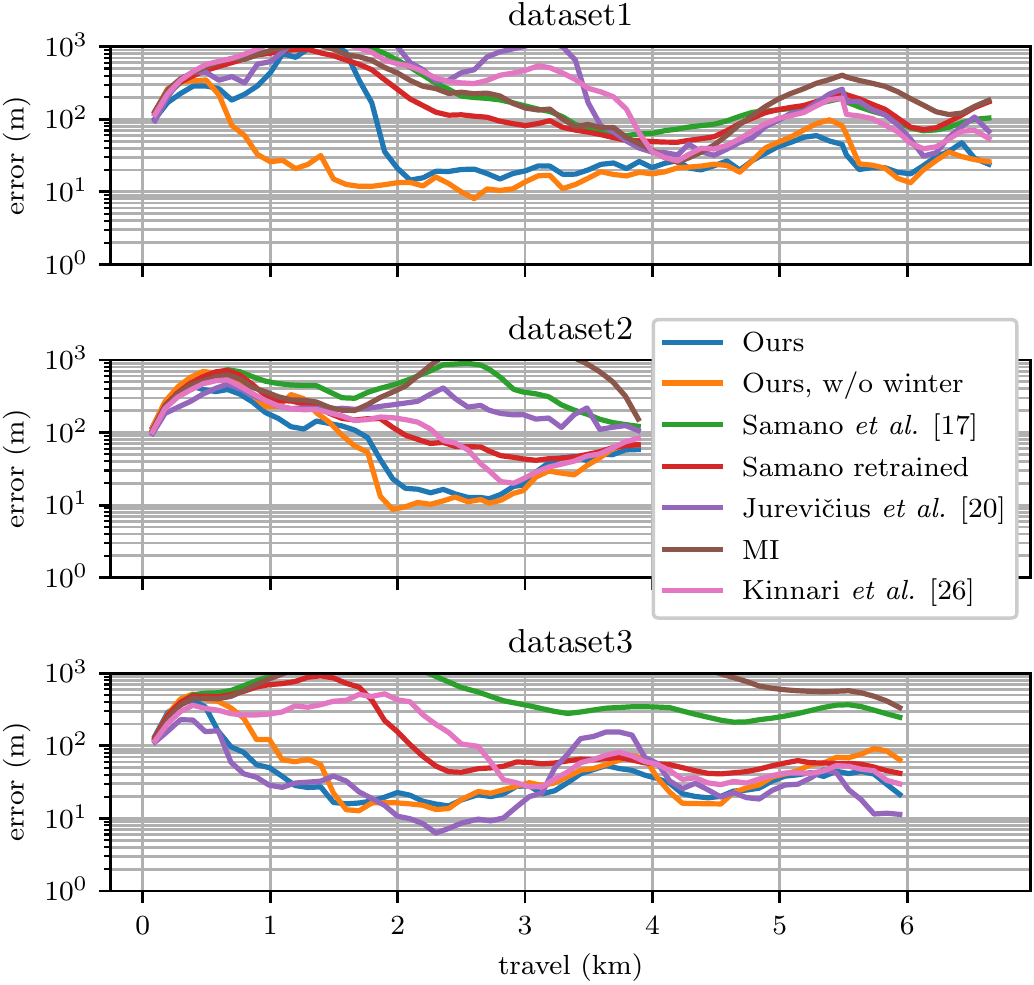}
    \caption{Errors with different trajectories on real \ac{uav} data. Logarithmic vertical scale.}
    \label{fig:sweexperiment_mean_error}%
    \figvspace{}
\end{figure}

The weighted mean error in \xycoords{} for all three \ac{uav} flights is visualized in \figref{fig:sweexperiment_mean_error}. On all \ac{uav} datasets, solution appears to converge after approximately 2 km of travel. The mean errors after 2 km of travel for \ac{uav} datasets 1, 2 and 3 were 26.5 m, 29.1 m and 30.6 m, respectively. The localization error and rate of convergence appears to mainly follow the conclusions drawn from the orthoimage experiment. Interestingly, our method without winter data appears to converge faster on dataset 1, possibly due to better fit between the environmental conditions of that particular flight data and the training data. With our method, time to convergence is longer than with the orthoimage experiment. Mean error before convergence (at approximately 1\ldots1.5 km of travel) appears to exceed the 5\%\ldots95\% interval estimated with the orthoimage experiment with all \ac{uav} datasets.

\begin{figure}
   \centering
   \subfloat[\centering \ground{} pose]{\includegraphics[width=0.47\linewidth]{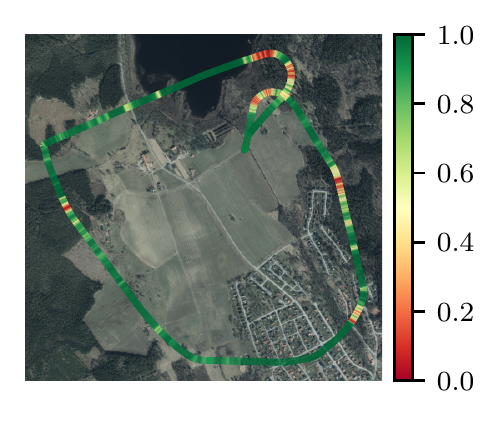}}%
   \subfloat[\centering random poses]{\includegraphics[width=0.47\linewidth]{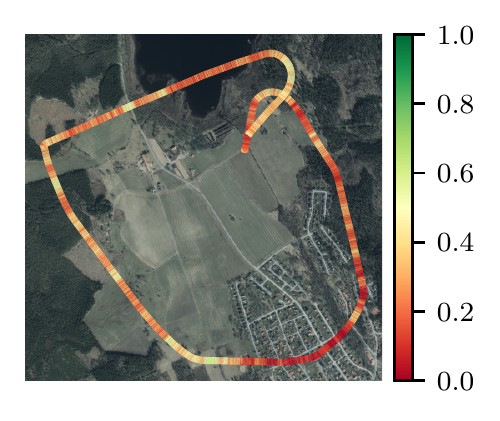}}%
    \caption{Value of similarity measure computed (a) with true poses of \ac{uav} dataset 3 and (b) as average of ten random poses per \ac{uav} frame, plotted over a map.}%
   \label{fig:similarity_measure_over_full_trajectory_uavds3}%
   \figvspace{}
\end{figure}

\begin{figure*}
\centering
\begin{tabularx}{\textwidth}{ cccccccc }%
\visualizeuavsample{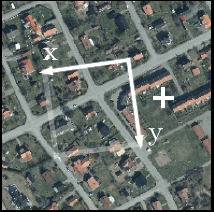}{Example 1}
\visualizeuavsample{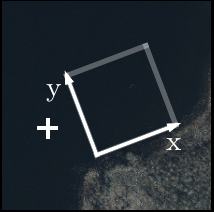}{Example 2}
\visualizeuavsample{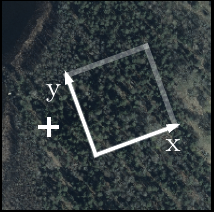}{Example 3}
 & \small (a) & \small (b) & \small (c) & \small (d) & \small (e) & \small (f) & \small(g) \\
\end{tabularx}
\caption{Similarity measure produced by the learning model under rotation, translation and scaling near true pose in example \ac{uav} image frames and poses. (a) $\mathcal{I}_t$ with outline of 96 m $\times$ 96 m square used for orthoprojecting $I_t$ highlighted; (b) $I_t$; (c) $I_{\mathcal{M},X_t^r}$ (here visualized using true pose as $X_t^r$); (d) true $X$ (white plus sign) and cropping square corresponding with $X_t^r$ overlaid on map; (e) similarity measure value when translating around true pose (translation along map coordinate axes); (f) and (g) similarity measure under rotations and scaling near true pose, respectively.}
\label{fig:uav_sample_pictures}
\figvspace{}
\end{figure*}

To better understand the performance of our similarity measure with the \ac{uav} datasets and understand the potential reason for this performance gap, we computed the similarity measure using \ground{} pose and plotted it for one of the \ac{uav} datasets in \figref{fig:similarity_measure_over_full_trajectory_uavds3}. For comparison, we also used the \ac{uav} image with ten random false poses per image to generate negative examples, and plotted their average similarity measure over the full trajectory. With an ideal method, the true poses would always yield a similarity measure very close to one, and the mean of random poses would be close to zero. From \figref{fig:similarity_measure_over_full_trajectory_uavds3} we see that similarity measure appears to be least reliable over forest areas. This failure mode appears in all of the \ac{uav} datasets and the geometric distortion of tall trees as seen from the \ac{uav} compared to the orthoview is a likely explanation to the initial difference in localization error performance between the \ac{uav} and orthoimage experiments.

Manual visual inspection of similarity measures produced by individual camera frames also leads to the remark that when the \ac{uav} is flying over a forest area where the density of trees is such that trees appear geometrically very distorted in the orthoprojection, similarity measure is often very low. Three exemplary \ac{uav} images, corresponding map crops, and similarity measure around true pose are visualized in \figref{fig:uav_sample_pictures}. Example 3 shows a case where geometric distortion of trees due to orthoprojection appears to affect similarity score significantly. Conversely, in an environment with numerous spatially distinct visual details such as buildings and roads (\figref{fig:uav_sample_pictures}, Example 1), the similarity measure shows a peak near the true state. The width of the peak is approximately 40 m in translation error and a few degrees in rotation error. Large width of the peak in proportion to the locality of the visual details is possibly affected by data augmentation, alignment errors of training data, or both. The network also does not appear to have high specificity; \eg{} in \figref{fig:uav_sample_pictures}, Example 1, adjacent intersections appear to produce high scores. The importance of texture is apparent also in \figref{fig:uav_sample_pictures}, Example 2:  there, for an image with extremely few visual features (taken when flying over a lake), the model produces a valid matching score in vicinity of the true state, but is unable to show a peaked output due to natural ambiguity of the environment. In such cases, our solution falls back to giving high likelihood for all particles corresponding to ambiguous terrain area, in effect relying on odometry, until unambiguous terrain areas are observed again.

Image patch extraction time is 0.33 s and inference time of $f$ is 0.13 s at each update on an Intel i7-9750H and NVidia Quadro RTX 3000 using $N=1000$ particles. We compute $f$ in batches of 100 image patches. Time consumption of both steps scales linearly with $N$ (\eg{} for $N=10 000$, times are 3.3 s and 1.3 s, respectively). As we perform an update every 100 m of travel, time between updates is significantly longer than the computing time for typical flight speeds of small \acp{uav}. This suggests the update can be run in real time, also on more resource constrained platforms. We exclude the analysis of computational requirements of \ac{vio} from this paper and refer the interested reader to \cite{8460664}.

\section{Discussion}
\label{sec:discussion}
The experiments with orthoimages demonstrate that by using a trained model for \ac{uav} image-to-map matching, significant reductions in convergence time and localization error can be achieved, compared to reference methods, in cases of both mild and significant seasonal appearance change in urban and non-urban environments. 

The comparison to Pearson correlation-based approach \cite{Jureviius2019RobustGL} and \ac{mi}-based approach hints towards the interpretation that in this task, trained models appear to outperform classical engineered methods in both convergence time and localization error, the only exceptions being the performance of \cite{9562005} on non-urban, significant appearance change, which is considerably out of training data domain in their solution, and the median error after convergence of \cite{Jureviius2019RobustGL} in the urban, minor appearance change case. In that one case, while \cite{Jureviius2019RobustGL} is able to achieve lower median error at convergence, we still achieve lower range of error across runs (\ie{} narrower 5th--95th percentiles) and faster convergence.

When looking at the comparison with the other learning method \cite{9562005}, our method performs better across all experiments. This can be partially explained by the fact that, in \cite{9562005}, the authors have not specifically trained their embedding generator to be robust to significant seasonal variance and they appear to focus on urban environments. This validates the need for season-invariant methods for visual localization of \acp{uav}. However, also in the case of in-domain data for their method (minor seasonal variation in urban environments), our method still provides faster convergence and smaller error. This, together with the slightly better localization performance of our model compared to the retrained Samano model, hints at the possibility that our method may be a more suitable for this task. 

The experiments with \ac{uav} data show that the model trained on orthoimages is able to localize also orthoprojected images from a \ac{uav} camera. The experiments also demonstrate that there is room for improvement on localization accuracy due to geometric appearance change introduced by an off-nadir viewpoint from a perspective camera.

Further investigation on model structure and dataset composition may yield improved results. The model was not trained specifically for a narrow peak of the output on correct pose; considering the training method and model structure may yield further improvements in localization, while incorporating a portion of \ac{uav} data in the training dataset might make the method more robust to perspective distortion.

\section{Conclusions}
\label{sec:conclusions}
We proposed a method for localizing a \ac{uav} with respect to an orthophoto map, in case of significant seasonal appearance change, trained using only satellite images taken at different times of year. We demonstrated the improvement in convergence time and localization error compared to \numrefmethods{} reference methods in simulated experiments involving significant seasonal appearance change. We showed the ability of the method to be used for localization of a real \ac{uav} and identified the most likely error sources for further development. 

Our results demonstrate that we can build models that are robust to appearance changes due to seasonal variations. However, seasons are only one source of variation for the dynamically changing operation environments of autonomous agents. The ability of agents to cope with all those variations will be crucial for their deployment in practice.

\bibliographystyle{./bibliography/IEEEtran}
\bibliography{./bibliography/IEEEabrv,./bibliography/bibliography}

\begin{thebibliography}{10}
\providecommand{\url}[1]{#1}
\csname url@samestyle\endcsname
\providecommand{\newblock}{\relax}
\providecommand{\bibinfo}[2]{#2}
\providecommand{\BIBentrySTDinterwordspacing}{\spaceskip=0pt\relax}
\providecommand{\BIBentryALTinterwordstretchfactor}{4}
\providecommand{\BIBentryALTinterwordspacing}{\spaceskip=\fontdimen2\font plus
\BIBentryALTinterwordstretchfactor\fontdimen3\font minus
  \fontdimen4\font\relax}
\providecommand{\BIBforeignlanguage}[2]{{%
\expandafter\ifx\csname l@#1\endcsname\relax
\typeout{** WARNING: IEEEtran.bst: No hyphenation pattern has been}%
\typeout{** loaded for the language `#1'. Using the pattern for}%
\typeout{** the default language instead.}%
\else
\language=\csname l@#1\endcsname
\fi
#2}}
\providecommand{\BIBdecl}{\relax}
\BIBdecl

\bibitem{Scaramuzza2020}
\BIBentryALTinterwordspacing
D.~Scaramuzza and Z.~Zhang, \emph{Aerial Robots, Visual-Inertial Odometry
  of}.\hskip 1em plus 0.5em minus 0.4em\relax Berlin, Heidelberg: Springer
  Berlin Heidelberg, 2020, pp. 1--9. [Online]. Available:
  \url{https://doi.org/10.1007/978-3-642-41610-1_71-1}
\BIBentrySTDinterwordspacing

\bibitem{7747236}
C.~{Cadena}, L.~{Carlone}, H.~{Carrillo}, Y.~{Latif}, D.~{Scaramuzza},
  J.~{Neira}, I.~{Reid}, and J.~J. {Leonard}, ``Past, present, and future of
  simultaneous localization and mapping: Toward the robust-perception age,''
  \emph{IEEE Transactions on Robotics}, vol.~32, no.~6, pp. 1309--1332, Dec
  2016.

\bibitem{ThrunProbabilisticRobotics05}
S.~Thrun, W.~Burgard, and D.~Fox, \emph{Probabilistic Robotics (Intelligent
  Robotics and Autonomous Agents)}.\hskip 1em plus 0.5em minus 0.4em\relax The
  MIT Press, 2005.

\bibitem{lowe2004distinctive}
D.~G. Lowe, ``Distinctive image features from scale-invariant keypoints,''
  \emph{International journal of computer vision}, vol.~60, no.~2, pp. 91--110,
  2004.

\bibitem{Cesetti2011AVG}
A.~Cesetti, E.~Frontoni, A.~Mancini, A.~Ascani, P.~Zingaretti, and S.~Longhi,
  ``A visual global positioning system for unmanned aerial vehicles used in
  photogrammetric applications,'' \emph{Journal of Intelligent \& Robotic
  Systems}, vol.~61, pp. 157--168, 2011.

\bibitem{MANTELLI2019304}
\BIBentryALTinterwordspacing
M.~Mantelli, D.~Pittol, R.~Neuland, A.~Ribacki, R.~Maffei, V.~Jorge,
  E.~Prestes, and M.~Kolberg, ``A novel measurement model based on abbrief for
  global localization of a uav over satellite images,'' \emph{Robotics and
  Autonomous Systems}, vol. 112, pp. 304--319, 2019.
\BIBentrySTDinterwordspacing

\bibitem{brief}
M.~Calonder, V.~Lepetit, C.~Strecha, and P.~Fua, ``Brief: Binary robust
  independent elementary features,'' in \emph{Computer Vision -- ECCV 2010},
  K.~Daniilidis, P.~Maragos, and N.~Paragios, Eds.\hskip 1em plus 0.5em minus
  0.4em\relax Berlin, Heidelberg: Springer Berlin Heidelberg, 2010, pp.
  778--792.

\bibitem{9311612}
H.~Hou, Q.~Xu, C.~Lan, W.~Lu, Y.~Zhang, Z.~Cui, and J.~Qin, ``Uav pose
  estimation in gnss-denied environment assisted by satellite imagery deep
  learning features,'' \emph{IEEE Access}, vol.~9, pp. 6358--6367, 2021.

\bibitem{Dusmanu_2019_CVPR}
M.~Dusmanu, I.~Rocco, T.~Pajdla, M.~Pollefeys, J.~Sivic, A.~Torii, and
  T.~Sattler, ``D2-net: A trainable cnn for joint description and detection of
  local features,'' in \emph{Proceedings of the IEEE/CVF Conference on Computer
  Vision and Pattern Recognition (CVPR)}, June 2019.

\bibitem{6126544}
E.~Rublee, V.~Rabaud, K.~Konolige, and G.~Bradski, ``Orb: An efficient
  alternative to sift or surf,'' in \emph{2011 International Conference on
  Computer Vision}, 2011, pp. 2564--2571.

\bibitem{Choi2020BRMLU}
J.~Choi and H.~Myung, ``Brm localization: Uav localization in gnss-denied
  environments based on matching of numerical map and uav images,'' \emph{2020
  IEEE/RSJ International Conference on Intelligent Robots and Systems (IROS)},
  pp. 4537--4544, 2020.

\bibitem{Dumble2015AirborneVN}
S.~J. Dumble and P.~Gibbens, ``Airborne vision-aided navigation using road
  intersection features,'' \emph{Journal of Intelligent \& Robotic Systems},
  vol.~78, pp. 185--204, 2015.

\bibitem{volkova2018more}
A.~Volkova and P.~W. Gibbens, ``More robust features for adaptive visual
  navigation of uavs in mixed environments,'' \emph{Journal of intelligent \&
  robotic systems}, vol.~90, no.~1, pp. 171--187, 2018.

\bibitem{Schleiss2019TRANSLATINGAI}
M.~Schleiss, ``Translating aerial images into street-map-like representations
  for visual self-localization of uavs,'' \emph{ISPRS - International Archives
  of the Photogrammetry, Remote Sensing and Spatial Information Sciences}, vol.
  4213, pp. 575--580, 2019.

\bibitem{masselli2016}
A.~Masselli, R.~Hanten, and A.~Zell, ``Localization of unmanned aerial vehicles
  using terrain classification from aerial images,'' in \emph{Intelligent
  Autonomous Systems 13}, E.~Menegatti, N.~Michael, K.~Berns, and H.~Yamaguchi,
  Eds.\hskip 1em plus 0.5em minus 0.4em\relax Cham: Springer International
  Publishing, 2016, pp. 831--842.

\bibitem{8575361}
A.~Nassar, K.~Amer, R.~ElHakim, and M.~ElHelw, ``A deep cnn-based framework for
  enhanced aerial imagery registration with applications to uav
  geolocalization,'' in \emph{2018 IEEE/CVF Conference on Computer Vision and
  Pattern Recognition Workshops (CVPRW)}, 2018, pp. 1594--159\,410.

\bibitem{9562005}
N.~Samano, M.~Zhou, and A.~Calway, ``Global aerial localisation using image and
  map embeddings,'' in \emph{2021 IEEE International Conference on Robotics and
  Automation (ICRA)}, 2021, pp. 5788--5794.

\bibitem{couturier2021}
\BIBentryALTinterwordspacing
A.~Couturier and M.~A. Akhloufi, ``{Convolutional neural networks and particle
  filter for UAV localization},'' in \emph{Unmanned Systems Technology XXIII},
  H.~G. Nguyen, P.~L. Muench, and B.~K. Skibba, Eds., vol. 11758, International
  Society for Optics and Photonics.\hskip 1em plus 0.5em minus 0.4em\relax
  SPIE, 2021, pp. 108 -- 120. [Online]. Available:
  \url{https://doi.org/10.1117/12.2585986}
\BIBentrySTDinterwordspacing

\bibitem{pearson1896vii}
K.~Pearson, ``Vii. mathematical contributions to the theory of
  evolution.—iii. regression, heredity, and panmixia,'' \emph{Philosophical
  Transactions of the Royal Society of London. Series A, containing papers of a
  mathematical or physical character}, no. 187, pp. 253--318, 1896.

\bibitem{Jureviius2019RobustGL}
R.~Jurevičius, V.~Marcinkevičius, and J.~Šeibokas, ``Robust gnss-denied
  localization for uav using particle filter and visual odometry,''
  \emph{Machine Vision and Applications}, vol.~30, pp. 1181 -- 1190, 2019.

\bibitem{Patel2020VisualLW}
B.~Patel, T.~D. Barfoot, and A.~P. Schoellig, ``Visual localization with google
  earth images for robust global pose estimation of uavs,'' in \emph{2020 IEEE
  International Conference on Robotics and Automation (ICRA)}, 2020, pp.
  6491--6497.

\bibitem{9357892}
M.~Bianchi and T.~D. Barfoot, ``Uav localization using autoencoded satellite
  images,'' \emph{IEEE Robotics and Automation Letters}, vol.~6, no.~2, pp.
  1761--1768, 2021.

\bibitem{6943040}
A.~Yol, B.~Delabarre, A.~Dame, J.-E. Dartois, and E.~Marchand, ``Vision-based
  absolute localization for unmanned aerial vehicles,'' in \emph{2014 IEEE/RSJ
  International Conference on Intelligent Robots and Systems}, 2014, pp.
  3429--3434.

\bibitem{8793558}
H.~Goforth and S.~Lucey, ``Gps-denied uav localization using pre-existing
  satellite imagery,'' in \emph{2019 International Conference on Robotics and
  Automation (ICRA)}, 2019, pp. 2974--2980.

\bibitem{9336674}
C.~Masone and B.~Caputo, ``A survey on deep visual place recognition,''
  \emph{IEEE Access}, vol.~9, pp. 19\,516--19\,547, 2021.

\bibitem{kinnari2021gnssdenied}
J.~Kinnari, F.~Verdoja, and V.~Kyrki, ``Gnss-denied geolocalization of uavs by
  visual matching of onboard camera images with orthophotos,'' in \emph{2021
  20th International Conference on Advanced Robotics (ICAR)}, 2021, pp.
  555--562.

\bibitem{5959226}
A.~Martinelli, ``Vision and imu data fusion: Closed-form solutions for
  attitude, speed, absolute scale, and bias determination,'' \emph{IEEE
  Transactions on Robotics}, vol.~28, no.~1, pp. 44--60, 2012.

\bibitem{Zagoruyko_2015_CVPR}
S.~Zagoruyko and N.~Komodakis, ``Learning to compare image patches via
  convolutional neural networks,'' in \emph{Proceedings of the IEEE Conference
  on Computer Vision and Pattern Recognition (CVPR)}, June 2015.

\bibitem{2018arXiv180906839B}
A.~Buslaev, A.~Parinov, V.~I.~I. E.~Khvedchenya, and A.~A. Kalinin,
  ``{Albumentations: fast and flexible image augmentations},'' \emph{ArXiv
  e-prints}, 2018.

\bibitem{scott1992}
D.~W. Scott, \emph{Multivariate Density Estimation}.\hskip 1em plus 0.5em minus
  0.4em\relax John Wiley \& Sons, Ltd, 1992.

\bibitem{8460664}
J.~Delmerico and D.~Scaramuzza, ``A benchmark comparison of monocular
  visual-inertial odometry algorithms for flying robots,'' in \emph{2018 IEEE
  International Conference on Robotics and Automation (ICRA)}, 2018, pp.
  2502--2509.

\end{thebibliography}

\end{document}